\documentclass[10pt,journal]{IEEEtran}
\pdfoutput=1
\usepackage{ifpdf}
\ifpdf

\ifCLASSOPTIONcompsoc
\usepackage[nocompress]{cite}
\else
\usepackage{cite}
\fi
\ifCLASSINFOpdf
\usepackage[pdftex]{graphicx}
\else
\fi
\usepackage{mathtools}
\usepackage{amsmath}
\usepackage{amssymb}
\usepackage{bbm}
\usepackage{booktabs}
\usepackage{epstopdf}
\DeclareMathOperator{\rank}{rank}

\usepackage{commath}
\usepackage{textcomp}
\usepackage{algorithm}
\usepackage[noend]{algpseudocode}

\makeatletter
\def\BState{\State\hskip-\ALG@thistlm}
\makeatother

\usepackage{array}

\ifCLASSOPTIONcompsoc
\usepackage[caption=false,font=footnotesize,labelfont=sf,textfont=sf]{subfig}
\else
\usepackage[caption=false,font=footnotesize]{subfig}
\fi

\ifCLASSOPTIONcaptionsoff
\usepackage[nomarkers]{endfloat}
\let\MYoriglatexcaption\caption
\renewcommand{\caption}[2][\relax]{\MYoriglatexcaption[#2]{#2}}
\fi
\usepackage{url}
\usepackage{mdwmath}
\usepackage{mdwtab}
\usepackage[table,xcdraw]{xcolor}
\usepackage{xcolor}
\usepackage{multirow}

\begin{document}
	%
	\title{On Clustering and Embedding Mixture Manifolds using a Low Rank Neighborhood Approach}
	%
	%
	%
	%
	
	\author{Arun~M.~Saranathan,~\IEEEmembership{Student Member,~IEEE,}
		and~Mario~Parente,~\IEEEmembership{Member,~IEEE}
		\thanks{A.~M.~Saranathan and M.~Parente are with the Department
			of Electrical and Computer Engineering, University of Massachusetts, Amherst, MA USA 01003 e-mail: (asaranat@umass.edu, mparente@ecs.umass.edu).}
		\thanks{This work was supported in part by the National Science Foundation under grant number IIS-1319585}
	}
	\maketitle

	
	\begin{abstract}
		Samples from intimate (non-linear) mixtures are generally modeled as being drawn from a smooth manifold. Scenarios where the data contains multiple intimate mixtures with some constituent materials in common can be thought of as manifolds which share a boundary. Two important steps in the processing of such data are (i) to identify (cluster) the different mixture-manifolds present in the data and (ii) to eliminate the non-linearities present the data by mapping each mixture-manifold into some low-dimensional euclidean space (embedding). Manifold clustering and embedding techniques appear to be an ideal tool for this task, but the present state-of-the-art algorithms perform poorly for hyperspectral data, particularly in the embedding task. We propose a novel reconstruction-based algorithm for improved clustering and embedding of mixture-manifolds. The algorithms attempts to reconstruct each target-point as an affine combination of its nearest neighbors with an additional rank penalty on the neighborhood to ensure that only neighbors on the same manifold as the target-point are used in the reconstruction. The reconstruction matrix generated by using this technique is block-diagonal and can be used for clustering (using spectral clustering) and embedding. The improved performance of the algorithms vis-a-vis its competitors is exhibited on a variety of simulated and real mixture datasets.
	\end{abstract}
	

	%

	\section{Introduction}\label{sec:introduction}

	%
	%
	%
	%
	Hyperspectral imagers (HSIs) measure electromagnetic energy scattered in their field of view in the Visible to Near InfraRed (VNIR) wavelength range (400-2500 nm) \cite{chang2007hyperspectral}.  HSI data-sets are organized into planes that form a data cube: each plane corresponds to solar electromagnetic energy reflected off the surface, acquired over a narrow wavelength range (spectral channel). Each pixel, now represents the vector of measurements acquired at a given location for all spectral channels -- a (reflectance) spectrum \cite{eismann2012hyperspectral}. Since each material is uniquely characterized by its (reflectance) spectrum \cite{liangrocapart1998mixed}, the spectra in an HSI can be used to identify the different materials present in the scene, making HSI an ideal tool for remote sensing of both the earth [e.g.  the Airborne Visual/Infrared Imaging Spectrometer (AVIRIS) \cite{green1998imaging}] and other planetary bodies [e.g. the Compact Reconnaissance Imaging Spectrometer for Mars (CRISM) \cite{murchie2007compact} and the Moon Mineralogy Mapper \cite{pieters2009moon}]. Since such sensors are generally deployed on airplane or satellite based platforms, the scene captured by these sensors covers a relatively large area on the ground (each CRISM pixel corresponds to approximately $18$ m on the ground; an AVIRIS pixel corresponds to approximately $20$ m in EO-$2$ mode). As a consequence the measured pixel spectrum is generally a combination (mixture) of the spectra of the constituent materials in the scene. The task of identifying the constituent materials (end-members) and their fractional abundances is referred to as the \textit{mixing/unmixing} problem \cite{keshava2002spectral}.
	
	If one assumes that each ray of light only interacts with single material (end-member) and that the mixing occurs at the sensor \cite{unmixingLinearReview}, then the mixed spectrum can be modeled as weighted (linear) combination of the end-member spectra. This model is a good approximation in the case of \textit{coarse checkerboard mixtures}. Images of a geological scenes on the other hand exhibit significant secondary scattering (due to the presence of \textit{nontrivial spatial structures} and/or \textit{microscopic} mixtures), in such a scenario the mixed spectrum will be a non-linear combination of the end-member spectra. Hyperspectral Images (HSI) of planetary/terrestrial surfaces typically cover areas occupied by multiple microscopic (intimate) mixtures. The mixing in these scenarios is modeled using radiative-transfer approaches, such as the one introduced by Hapke \cite{hapkeBook}. In \cite{Saranathan14ManifoldClustering}, it has been shown that the point cloud representing intimately mixed spectra of known materials (endmembers), if modeled using Hapke's model can be considered as lying near a manifold obtained by sampling an \emph{abundance} simplex,  i.e. samples drawn from a mixture with $N$ endmembers can be modeled as a non-linear mapping of an $N-1$ dimensional simplex. If the data contains more that one such mixtures the data can be modeled as lying on different mixture-manifolds.
	
	Unmixing of such data requires identifying the different mixtures (manifold clustering), and retrieving the abundance simplex for each mixture-manifold, accomplished by an embedding into a lower dimensional euclidean space -- such processing leads to significant gains in terms of both future processing (e.g. unmixing) and data storage. 
	In the absence of shared endmembers the mixture-manifolds appear well separated, and simply using a clustering technique such as N-Cuts \cite{Shi:2000:NCI:351581.351611} is sufficient to identify the different mixture-manifolds, following which accurate low dimensional parameter space representations (embeddings) for each of the mixture-manifolds can be computed using manifold learning \cite{huo2007survey}. On the other hand, if the different manifolds overlap, identifying the different manifolds and generating quality embedding is not as easily accomplished.
	
	In the manifold learning literature there exists a variety of techniques for identifying the different manifolds present the data. The simplest approaches model the data as lying on or near linear manifolds (affine spaces) and leverages this expectation of linearity to aid manifold identification (see \cite{subspaceClustering} and references therein). These methods learn a similarity matrix whose entries measures similarities among the data-points, in addition the algorithms include some constraints or penalties to ensure that the similarity matrix is block diagonal. The Low Rank Representation (LRR) \cite{LRR} which attempts to generate a ``lowest rank representation among the candidates to represent the data". Similarly the Sparse Subspace Clustering (SSC) \cite{elhamifar2013sparse} attempts to find the ``sparsest representation" of the data. The Correlation Adaptive Subspace Segmentation (CASS) \cite{lu2013correlation} uses the \textit{trace Lasso} norm \cite{grave2011trace} to group correlated data together. Most of these algorithms assume independent subspaces \footnote{A group of subspaces $\mathcal{U}_i$ are considered independent if $\mathcal{U}_i~\cap~\mathcal{U}j|_{j \neq i}$ = \{0\} and $dim(\mathcal{U}_i \cup (\cup_{j \neq i} \mathcal{U}_j))  dim(\mathcal{U}_i)+dim (\cup_{j \neq i} \mathcal{U}_j)) $}, a more recent work \cite{peng2016constructing} shows that property of \textit{Intrasubspace Projection Dominance (IPD)}, i.e. only points on the same subspace as the target point are assigned high affinities, holds even for dependent subspaces under certain constraints on the data distribution for the $\ell_{1},~\ell_2$ and $\ell_{\infty}$ norms. Other algorithms such as the Local Structural Consistency (LSC) \cite{wang2011local} and Spectral Multi-Manifold clustering (SMMC) create a structure-dependent similarity metric to generate suitable affinity matrices for spectral clustering. Another approach, the Robust Multiple Manifolds Structure Learning (RMMSL) \cite{gong2012robust}, generates affinity matrices based on tangent space alignment. Unfortunately, these algorithms do not generate low-dimensional embedding for the data. 
	
	Manifold Clustering \& Embedding (MCE) algorithms are capable of simultaneously classifying and embedding the data. Some algorithms like k-Manifolds \cite{Souvenir05manifoldclustering} are based on the assumption that there is no embedding in a lower dimensional space that preserves all the properties captured in the high dimensional data, but this has been shown to be false in the case of nonlinear manifolds that only share a boundary, such as mixture-manifolds with shared endmembers\cite{Saranathan14ManifoldClustering}. Other popular MCE algorithms are based on the notion of reconstruction coefficients/matrices first introduced in Locally Linear Embedding (LLE) \cite{LLE}. The algorithms follow a general scheme wherein each data-point is expressed as an affine (or sometimes linear) combination of other points in the data-set. In addition to this, the algorithms place some penalty on the reconstruction coefficients to make sure that data-points on the same manifold are ``chosen" (assigned non-zero reconstruction coefficients). This yields a reconstruction matrix that is block-diagonal and the application of a spectral clustering algorithm to such a matrix is sufficient to identify the different manifolds present in the data. The reconstruction matrices can also be used to generate embeddings like in the LLE.
	
	Reconstruction-based methods differ mainly in the type of constraints imposed to generate a block-diagonal reconstruction matrix. One such algorithms is the Low Rank Embedding (LRE) \cite{Liu:2011}, the algorithm reconstructs each point as a linear combination of the other data-points, and adds a rank penalty on the reconstruction matrix to ensure that it is block-diagonal. In effect the LRE assumes that the reconstruction coefficients of data-points on the same manifold have a ``similar underlying structure", i.e. they can be reconstructed accurately by using the same set of points. The LRE algorithm generates an embedding of the data into a low-dimensional space using the reconstruction matrix in the same fashion as the LLE. The algorithm then performs k-means \cite{kMeans} on the embedding to learn manifold memberships. While the assumption on the structure of reconstruction coefficients is a reasonable assumption for data which can be modeled as being drawn linear subspaces with some distortions, in scenarios where there are highly nonlinear manifolds different sets of points are prioritized to reconstruct target points in different parts of the manifold. Furthermore, the embedding scheme used only captures the geometric properties of a neighborhood when the reconstruction coefficients are unaffected by translation, rotation and scaling \cite{LLE}. The LLE algorithm ensures such invariance by enforcing a \textit{sum-to-one} (affineness) constraint on the reconstruction coefficients in the neighborhood. Since LRE does not include this constraint, the reconstruction coefficients are no longer invariant to rigid linear transformations which leads to distortions in the embedding.
	
	Another example of a reconstruction-matrix based MCE algorithm is the Sparse Manifold Clustering and Embedding (SMCE) \cite{NIPS2011_4246}. The SMCE attempts to find a reconstruction matrix where each data-point is expressed as an affine combination of its $k$-nearest neighbors and adds an additional penalty on the distance based-sparsity of the reconstruction coefficient vector. The authors show that the effect of minimizing both reconstruction error and sparsity penalty as much as possible is that only data-points on the same manifold are assigned non-zero weights. Creating \emph{sparse} neighborhoods aids the clustering objective, it throws up some issues in the embedding. In particular, the spectral embedding technique introduced in the LLE  only preserves local relationships, and there is no penalty if the global geometric information is distorted. Namely, if different neighborhoods do not share points, there is no penalty if they are embedded with different scalings or rotations: therefore global shape is only preserved if there is significant overlap between adjacent neighborhoods. Since the SMCE creates very sparse neighborhoods with little or no overlap leading to significant distortions in the global shape. 
	
	The Bundle Manifold Embedding (BME) \cite{li2010learning} instead constructs two graphs - one to capture the structure of the local neighborhood (this graph is similar to the neighborhood graph in Laplacian Eigenmaps \cite{Belkin01laplacianeigenmaps}) and another thresholded reconstruction graph to illustrate manifold memberships. The reconstruction graph is similar to the one constructed in LLE except that each row is now thresholded to preserve only the $d+1$ largest magnitudes (where $d$ is the intrinsic dimension of the manifold). The final neighborhood graph is convex combination of these two neighborhood graphs. If the first graph is weighted heavily it would generate a faithful embedding but the classification performance is not guaranteed, on the other hand if the second graph is weighted heavily the classification is expected to be better but the embedding suffers (as the assumptions of neither the LLE or LE hold for this matrix). A more recent technique, the Joint Manifold Clustering and Embedding (JMCE) \cite{JMCE}, expresses each data-point as a convex combination of its $k$ nearest neighbors and at the same time adds a penalty on the magnitude of the non-zero weights assigned to neighbors on other manifolds. While the technique has shown some promise in clustering of hyperspectral data, due to the restriction to convex reconstructions, the embedding suffers from distortions at the boundary of the manifolds. 
	
	In this paper we propose a novel reconstruction based approach, the Low Rank Neighborhood Embedding (LRNE), which  expresses every data-point as an affine combination of its $k$ nearest neighbors, in addition we also add a \textit{trace-Lasso} \cite{grave2011trace} norm penalty on the weighted neighborhood to ensure that only neighbors on the same manifold are prioritized for the reconstruction. Specifically, the \textit{trace-Lasso} penalty prevents neighborhood-points from different mixture-manifolds being assigned large weights thus ensuring block-diagonality in the reconstruction matrix. Since the reconstruction scheme is local and affine, the LRNE reconstruction matrix can be embedded by using a spectral embedding stage. Also, since the LRNE does not prioritize  sparsity as much as the SMCE, the neighborhoods show sufficient overlap between adjacent neighborhoods for better embedding. A preliminary version of this algorithm with limited results can be found in \cite{Saranathan16ManifoldClustering}. 
	
	The paper is arranged as follows: in section \ref{algorithm} we describe the new Low Rank Neighborhood Embedding (LRNE), we will provide both intuition and theoretical basis to show that choosing a low-dimensional neighborhood will ensure that only data-points from the same manifold are chosen. Following this we will describe an optimization scheme that will ensure the choice of such a low-dimensional neighborhood and the steps required to generate the clustering and embedding from the reconstruction coefficients. In section \ref{experiments} we describe the experiments used to compare the various MCE algorithms and analysis of the results. We will offer concluding remarks and avenues for further research in section \ref{conclusion}.
	
	\begin{figure}[!t]
		\centering
		\includegraphics[width=0.35\textwidth]{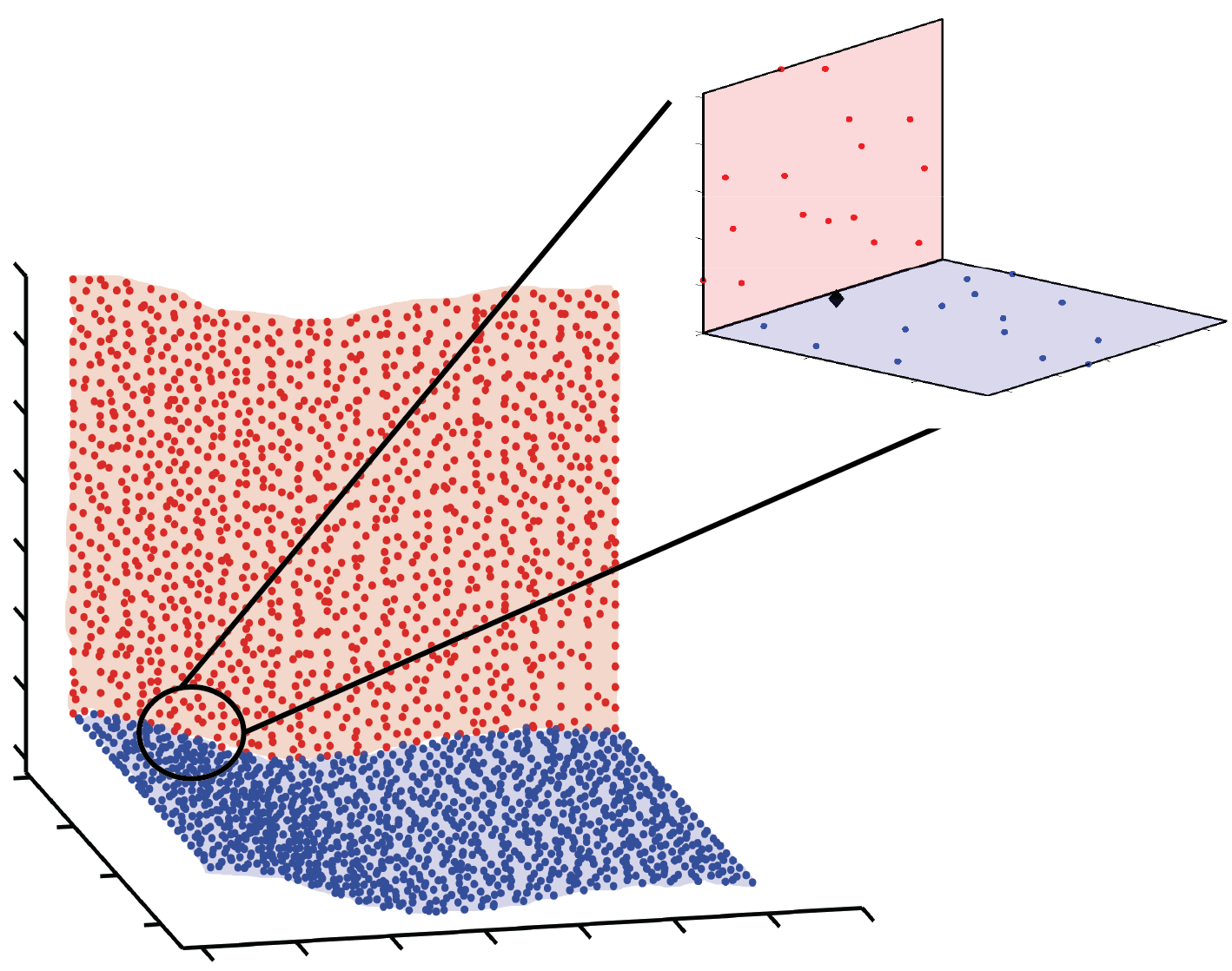}
		\caption{The neighborhood for a point at/near the intersection of manifolds}
		\label{fig_manifTypes}
	\end{figure}
	
	\section{The Low-rank Neighborhood Embedding Algorithm}
	\label{algorithm}
	
	\subsection{The Main Idea}
	Given a set of points in $\mathbb{R}^D$, drawn from $p$ different smooth and sufficiently well sampled manifolds, the LLE attempts to find the appropriate reconstruction coefficients, as \cite{LLE}:
	\begin{equation}
	\begin{aligned}
	& \underset{ \alpha}{\text{minimize}}
	& & ||x - \displaystyle\sum_{i=1}^{k} \alpha_i n_{i}||^{2} \\
	& \text{subject to}
	& & \mathbbm{1}^{T}\alpha=1.
	\end{aligned}
	\label{eqn:LLE}
	\end{equation}
	where $x$ is the target point for which we are finding the reconstruction coefficients, $|| \cdot ||$ is the $\ell_2$ norm, $N$ is the set of the $k$ nearest neighbors of the data-point $x$ defined by $\mathcal{N}(x)=\{n_1, n_2, \ldots n_k\}$ and $\alpha_i \in \mathbb{R}$ is the reconstruction coefficient assigned to the neighbor $n_i$. In scenarios where the data lie on multiple manifolds with some overlap, the neighborhood $\mathcal{N}(x)$ of a target point at/near an intersection will contain points from all the different manifolds that overlap at the intersection. Since each manifold is smooth and well sampled, the neighbors drawn from each manifold will appear to lie on a linear patch. The inset pictures in Fig. \ref{fig_manifTypes} show the neighborhoods for a data-point at/near the intersection for the example manifolds.
	
	The LRNE attempts to select from the neighborhood $\mathcal{N}(x)$ only those neighbors that lie on the same manifold as the target point for the reconstruction. This problem can be cast as the task of finding a subset of the neighborhood that both successfully reconstruct the target point (according to Eqn. \eqref{eqn:LLE}) and spans an affine space of the lowest possible dimension. Consider a weighted neighborhood matrix $ M = [ \alpha_1 n_1 ~ \alpha_2 n_2 \ldots ~\alpha_{k} n_{k}]$. If we add a penalty to Eqn.  \eqref{eqn:LLE} based on the dimension of the space spanned by the columns of $M$ (which is given by $\rank(M)$), the only way to lower the dimensionality of such space (i.e. $\rank(M)$) is by zeroing out coefficients $\alpha_i$, corresponding to all the points that lie on some of the manifolds. We informally refer to the neighbors with non-zero coefficients as the points ``chosen" for the reconstruction. The penalty on the dimensionality of the chosen set can be increased so that only neighbors lying on one manifold are ``chosen". 
	
	Note that, if some of the chosen points do not lie on the same manifold as the target, the reconstruction error will not be minimized. Thus the need to simultaneously minimize the reconstruction error and the dimension of the neighborhood will ensure that only points on the same manifold as the target point are ``chosen"(in the absence of noise). The parameter $\lambda$ regulates the trade-off between allowing reconstruction using some points from the wrong manifold to obtain a better linear fit to the data when noise or local density changes are present. As a result of the above discussion, we propose the addition of the following dimension based penalty to the LLE reconstruction objective:
	\begin{equation}
	\begin{aligned}
	& \underset{ \alpha}{\text{minimize}}
	& & \dfrac{1}{2} ||x - \displaystyle\sum_{i=1}^{k} \alpha_i n_{i}||^{2} + \lambda~ \rank{(M)} \\
	& \text{subject to}
	& & \mathbbm{1}^{T}\alpha=1.
	\end{aligned}
	\label{eqn:solRank}
	\end{equation}
	It is straightforward to note that for target points not close to the intersection, Eqn. \eqref{eqn:solRank} can be  applied as well with the effect that no points from $\mathcal{N}(x)$ are excluded. The one hurdle in the solution of the problem defined in Eqn. \eqref{eqn:solRank} is that the $\rank$ function and hence the objective function defined above is not convex. To mitigate this problem the $\rank$ function is replaced by the \emph{nuclear norm} $(||\cdot||_{*})$ function which has been shown to be a convex approximation for the $\rank$ function \cite{Candes:2009:EMC:1666896.1666899}. The modified objective function can be written as:
	\begin{equation}
	\begin{aligned}
	& \underset{ \alpha}{\text{minimize}}
	& & \dfrac{1}{2} ||x - \displaystyle\sum_{i=1}^{k} \alpha_i n_{i}||^{2} + \lambda~ ||\hat{N} ~diag(\alpha)||_{*} \\
	& \text{subject to}
	& & \mathbbm{1}^{T}\alpha=1.
	\end{aligned}
	\label{eqn:lowRankNeigh_weights}
	\end{equation}
	where $\hat{N} = [\hat{n}_1~ \hat{n}_2 \ldots  \hat{n}_k]$, and $\hat{n}_i=(n_i -mean(\mathcal{N}_i))/||(n_i -mean(\mathcal{N}_i))||$. The normalization is necessary because unlike the $\rank$, the nuclear norm is affected by the scaling of the columns of $M$ or simple translation of the points. The result of application of Eqn.~\eqref{eqn:lowRankNeigh_weights} to the $i$-th target point in the data-set is a reconstruction vector $\alpha$ that fills the $i$-th row of a reconstruction matrix $R$ so that $R_{ij}=0$ if $x_j\notin \mathcal{N}(x_i)$, otherwise the value $R_{ij}$ is $\alpha_j$. 
	
	The norm-penalty used above, i.e. $||\hat{N} ~.diag(\alpha)||_{*}$  is known as the \textit{traceLasso} norm, $\ell_{\Omega_D}$ and was introduced in \cite{grave2011trace}. The \textit{traceLasso} norm interpolates between the $\ell_{1}$ and the $\ell_{2}$ norms depending on the correlations between the columns of the matrix $\hat{M}$. Since points belonging to the same manifold are drawn from the same affine space as the target point they have higher correlation to the target point as opposed to points drawn from the other spaces, therefore the penalty is expected to approximate an $\ell_{2}$ penalty for points on the same manifold while behaving as an $\ell_{1}$ penalty for neighbors on the other manifolds, i.e. in effect assigning large reconstruction coefficients for points on the other manifold carries a larger penalty thus forcing only neighbors on the same manifold to be ``chosen".
	
	In the following subsection we will show that if certain conditions on data distributions are met then only points on the same manifold as the target point are used for the reconstruction.
	\subsection{The Theoretical Background}
	If $x \neq 0$ is a data-point drawn from a union of subspaces that is spanned by $D=~[D_{x} ~D_{-x}]$, where $D_{x}$ and $D_{-x}$ are the set of points from the same subspace and other subspace as $x$. Let $\mathcal{S}_{D_{x}}$ and $\mathcal{S}_{D_{-x}}$ be the subspaces spanned by $D_{x}$ and $D_{-x}$ respectively.  Let $c^{*}$ be the optimal solution of:
	\begin{equation}
	min \|c\|_{p} ~~~s.t.~~~~x = Dc
	\end{equation}
	where $\|.\|_{p}$ refers to a general norm. Again without loss of generality we can let partition $c^{*}$ as $c^{*} = [c^{*}_{D_{x}}~~ c^{*}_{D_{-x}}]^T$, then it was shown in \cite{peng2016constructing} that:
	
	\textit{Lemma 1:} Let $y \in \mathcal{S}_{D_{x}}$ and $\hat{y} \in \mathcal{S}_{D_{-x}}$ be any two data-points belonging to the different subspaces. Then $z_{D_{x}}$ and $z_{D_{-x}}$, respectively, be the solutions of $min\|z\|_{p}~~s.t.~~ y=D_{x}z$  and $min\|z\|_{p}~~s.t.~~ \hat{y}=D_{-x}z$. If we now define the point $y=x~-~D_{x}c^{*}_{D_{x}} = D_{-x}c^{*}_{D_{-x}}$, the point $y$ can now be said to belong to $\mathcal{S}_{D_{x}} \cap \mathcal{S}_{D_{-x}}$. Then we have that $c^{*}_{D_{-x}}=0$ if and only if $\|z_{D_{x}}\|_p~<~\|z_{D_{-x}}\|_p$.\\
	\textit{Proof}: The proof for Lemma 1 can be seen in section III of \cite{peng2016constructing}.
	
	\textit{Lemma 2:} For a point $y$ defined as in Lemma 1, we will have that $\|z_{D_{x}}\|_p~<~\|z_{D_{-x}}\|_p$ if
	\begin{equation}
	\sigma_{min}(D_{x})~ \geq~ k_{p}~ cos(\theta_{min})~ \|D_{x}\|_{max,2} 
	\end{equation}
	where $\sigma_{min}(D_{x})$ is the smallest singular value of $D_{x}$, $\theta_{min}$ is the first principal angle between $D_{x}$ and $D_{-x}$, and $\|D_{x}\|_{max,2}$ is the maximum $\ell_{2}$ of the columns of $D_{-x}$.\\
	\textit{Proof}: A version of this proof only for the norms $p \in ~{1,2,\infty}$ was shown in \cite{peng2016constructing}. We will reproduce this proof with modifications to include the \textit{traceLasso norm} $\ell_{\Omega_D}$ in Appendix \ref{sec:ProofLemma2}.
	
	$\sigma{_{min}}(D_x)$ provides the estimation of power in the dimension of lowest variance. In the case of dependent subspaces the first principal angle $\theta_{min}=0$, and therefore there must either be sufficient variation in the dimension of lowest variance or a large number of samples in $D_x$ to ensure that the above conditions are met. Since in the case of manifold each patch can be modeled as being drawn from an affine spaces as opposed to subspaces we use the technique suggested in \cite{elhamifar2013sparse} and reconstruct each points as an affine combination of its neighbors rather than a linear combination of their neighbors.
	\subsection{Numerical Solution}
	In this section we will describe an optimization scheme to solve Eqn. \eqref{eqn:lowRankNeigh_weights}. The equation can be rewritten as: 
	\begin{equation}
	\begin{aligned}
	& \underset{\alpha}{\text{minimize}}
	& & \dfrac{1}{2} \norm{x - N\alpha}{^{2}} + \lambda \norm{\hat{N} \left(\sum_{i=1}^{k}\alpha_i e_{i} e_{i}^{T}\right)}_{*} \\
	& \text{subject to}
	& & \mathbbm{1}^{T} \alpha = 1
	\end{aligned}
	\label{eqn:fullOptimProblem}
	\end{equation}
	where the vector $e_i$ is a vector such that the $i$-th element is $1$ and all the other elements are $0$, where $\hat{N}$ is the normalized neighborhood matrix described in section \ref{algorithm}. All the terms in Eqn. \ref{eqn:fullOptimProblem} are convex and can be minimized using Alternating Direction Method of Multipliers (ADMM) \cite{ADMM}. Using a dummy variable the optimization problem can be written as:
	\begin{equation*}
	\begin{aligned}
	& \underset{\alpha,~V}{\text{minimize}}
	& & \dfrac{1}{2} \norm{x - N\alpha}{^{2}} + \lambda \norm{V}_{*} \\
	& \text{subject to}
	& & \mathbbm{1}^{T} \alpha = 1\\
	& & & V~=~\hat{N} \left({\sum_{i=1}^{k}}\alpha_i e_{i} e_{i}^{T} \right)
	\end{aligned}
	\end{equation*}
	The augmented Lagrangian for the ADMM can be written as:  
	\begin{multline}
	\mathcal{L}(V,\alpha,\Lambda_1,\Lambda_2) = \dfrac{1}{2} \norm{x - N\alpha}{^{2}} + \lambda ~\norm{V}_{*} \\ + \dfrac{\beta}{2}{||\mathbbm{1}^{T} \alpha - 1 + \dfrac{1}{\beta} \Lambda_1||}^{2}_{F} + \frac{\beta}{2}||V-~\hat{N} \sum_{i=1}^{k}\alpha_i e_{i} e_{i}^{T}  + \dfrac{1}{\beta} \Lambda_2||^{2}_{F}
	\label{eqn:fullLagrangian}
	\end{multline}
	The ADMM decomposes Eqn. \ref{eqn:fullLagrangian} into two separate optimization problems. 
	\subsubsection*{Update equation for \textbf{$V$}}
	\begin{equation*}
	V^{k+1} = \underset{V}{\text{argmin}}~~ \mathcal{L}(V,\alpha^k,\Lambda^{k}_{1},\Lambda^{k}_{2})
	\end{equation*}
	\begin{equation}
	V^{k+1} = \underset{V}{\text{argmin}} ~~\lambda~ \norm{V}_{*} + \dfrac{\beta}{2}\norm{V~-T+~\dfrac{1}{\beta}\Lambda^{k}_{2}}_{F}^{2}
	\label{eqn:minimizeVEqn}
	\end{equation}
	where $T = \hat{N}~\sum_{i=1}^{k}\alpha_i^k e_{i} e_{i}^{T}$. This equation is similar to Eqn. (6) in Lin et al. \cite{lin2010augmented} and  can be updated by using a proximal update using the singular value thresholding \cite{cai2010singular}.
	\subsubsection*{Update equation for \textbf{$\alpha$}}
	The next step is the optimization with respect to $\alpha$, which can be written as: -
	\begin{align}
	\alpha^{k+1} = & \underset{\alpha}{\text{argmin}} ~~\mathcal{L}(V^{k+1},\alpha,\Lambda^{k}_{1},\Lambda^{k}_{2})\nonumber \\
	= & \underset{\alpha}{\text{argmin}}~~\dfrac{1}{2} {\norm{x-N\alpha}{^{2}}} + \dfrac{\beta}{2} {\norm{\mathbbm{1}^{T} \alpha - 1 + \dfrac{1}{\beta} \Lambda_1^k}_{F}^{2}}+ \nonumber  \\
	& ~~~~~~~~~~~~~ + \dfrac{\beta}{2}{ \norm{ V^{k+1} - \hat{N}~\displaystyle\sum_{i=1}^{k}\alpha_i e_{i} e_{i}^{T} +\dfrac{1}{\beta} \Lambda_2^k}_{F}^{2}}
	\label{eqn:minimizeAlphaEqn}
	\end{align}
	The gradient of the above equation can be written as:
	\begin{equation}
	- N^Tx + N^T N \alpha - \beta \mathbbm{1}F_1 + \beta\mathbbm{1} \mathbbm{1}^{T} \alpha -\beta P + \beta Q \alpha = 0
	\end{equation}
	Solving the above equation we get that the optimal value is given by the stationary point
	\begin{equation}
	{\alpha}^{k+1} = \left( N^T N + \beta\mathbbm{1} \mathbbm{1}^{T} + Q \right)^{-1} \left( N^Tx + \beta \mathbbm{1}F_1 + \beta P \right)
	\label{eqn:alphaUpdate}
	\end{equation}
	\subsubsection*{Update for the Lagrangian Multipliers}
	The Lagrangian multipliers can the be updated as:-
	\begin{equation}
	\Lambda_1^{k+1} = \Lambda_1^{k} + \beta (\mathbbm{1}^{T} \alpha - 1)
	\label{eqn:LagMult_1}
	\end{equation}
	\begin{equation}
	\Lambda_2^{k+1} = \Lambda_2^{k} + \beta (V - \hat{N} \displaystyle\sum_{i=1}^{k}\alpha_i e_{i} e_{i}^{T} ) 
	\label{eqn:LagMult_2}
	\end{equation}
	The full Low Rank Neighborhood Embedding Scheme is shown as Algorithm \eqref{alg:LRNE}.
	
	\begin{algorithm}
		\caption{The Low Rank Neighborhood Embedding}\label{alg:LRNE}
		\begin{algorithmic}[1]
			\State \textbf{Input:}	A set of $n$ points $X = [x_1,~x2,...,x_n] \in \mathbb{R}^{d}$, the number of neighbors $k$ and the trade of parameter $\lambda$
			\State \textbf{Initialize:} A reconstruction matrix $R \in \mathbb{R}^{n \times n}$, s.t. $R_{ij}=0,~\forall i,j$
			\For{every point in X}
			\State Find $N$ = $k-$nearest neighbors of the point $x_i$	
			\While{not converged}
			\State Solve $V^{k+1} = \underset{V}{\text{argmin}}~~ \mathcal{L}(V,\alpha^k,\Lambda^{k}_{1},\Lambda^{k}_{2})$ using the SVT as per \cite{lin2010augmented}
			\State Solve for $\alpha^{k+1}$ using Eqn. \eqref{eqn:alphaUpdate}
			\State Update the Lagrangian multipliers using Eqns. \eqref{eqn:LagMult_1} \& \eqref{eqn:LagMult_2}		
			\EndWhile
			\For{$j=1$ to $n$}
			\If{($x_j \in \mathcal{N}(x_i)$)}
			\State $R(i,j) = \alpha_j $	
			\EndIf
			
			\EndFor
			\EndFor
			\State \textbf{Output:} The reconstruction matrix $R$.
		\end{algorithmic}
	\end{algorithm}
	\subsection{Clustering from the Reconstruction Matrix}
	Based on the discussion in the previous section, the reconstruction matrices will be approximately block diagonal with the blocks corresponding to different mixture-manifolds. The set of data-points are modeled as the nodes in a graph where the similarity between the nodes $x_i$ and $x_j$ is based on the reconstruction coefficient $(r_{ij})$ in $R$ and the distance between the nodes. The similarity between the two nodes is defined as $w_{ij} = \frac{r_{ij}/||x_j-x_i||_{2}}{\sum_{t \ne i} r_{it}/||x_t-x_i||_{2}}$ (similar to the one defined in section $2.2$ of \cite{NIPS2011_4246}). The final step is to make this similarity matrix symmetric, this can be achieved by setting $W_{sym} = \max(W , W^T)$. This matrix is then  provided as an input to a spectral clustering algorithm\cite{Shi:2000:NCI:351581.351611,yang2012clustering}.
	\subsection{Embedding from the Reconstruction Matrix} 
	Since the LRNE objective function includes an affiness constraint, the coefficients are unaffected by translations, rotation or scalings of the data which ensures that these coefficients capture the ``intrinsic geometric structure" of the neighborhood \cite{LLE}. 
	Therefore, we find a low-dimensional representation $Y$ solving the following problem : 
	\begin{equation}
	\begin{aligned}
	& \underset{Y}{\text{minimize}}
	& & ||Y - YR||{^2_F}  \\
	& \text{subject to}
	& & YY{^T} = I.
	\end{aligned}
	\label{embedEquation}
	\end{equation}
	Minimizing Eqn. \ref{embedEquation} generates the coordinates of the low-dimensional embedding $Y$, centered at the origin \cite{LLE}. Following this the embeddings corresponding to each of the different classes are separated based on the classification to generate individual manifold embeddings.	
	\subsection{A note on Time-Complexity}
	The CVX solver uses interior point techniques, which are efficient but slow, and have a worst case time complexity of the order of $\mathcal{O}(n^6)$ \cite{LADMAP}. Each iteration of the ADMM algorithm has a time complexity of  $\mathcal{O}(mn^2)$  since it requires the Singular Value Decomposition (SVD) of a $m \times n$ matrix\cite{golub2012matrix}. In comparison the SMCE has a time complexity of  $\mathcal{O}(k)$ (where k is number of iterations for the optimization) . The LRE has a time complexity of $\mathcal{O}(N^3)$, where $N$ is the number of points but is in general faster than the LRNE as it solves for the reconstruction coefficients of all the data-points at once. In the future we will look to adapt techniques such as the one described in \cite{LADMAP}, which have been previously used to further improve the time complexity of rank-based techniques.
	
	\section{Experiments and Results}
	\label{experiments}
	The new algorithm was tested on some simulated \& real (benchmark) hyperspectral mixture data-sets for MCE. The performance of the LRNE is compared with the BME, LRE and SMCE algorithms. The LRNE outperformed its competitors in terms of both classification and embedding.
	\subsection{Datasets \& Experiments Setup} \label{sec:datasets}
	\subsubsection{The Simulated Mixtures Dataset}
	This dataset simulates the scenario in which a hyperspectral imager observes several pixels from a terrain composed \emph{intimate} mixtures of different  {endmembers} , as in a sand beach made up of grains of different  minerals. 
	The mixing is modeled using the Hapke \cite{hapkeBook} mixing model. It has been shown in \cite{Saranathan14ManifoldClustering} that even if the abundance simplex is uniformly sampled, the nonlinear Hapke mapping produces a point cloud that exhibits a density gradient, with higher density (of samples) near the dark endmembers and lower density around the brighter endmembers. The exact nature and amount of the density gradient depends upon the endmembers chosen in the mixture. 
	
\begin{figure*}[!t]
		\centering
		\includegraphics[width=0.65\paperwidth]{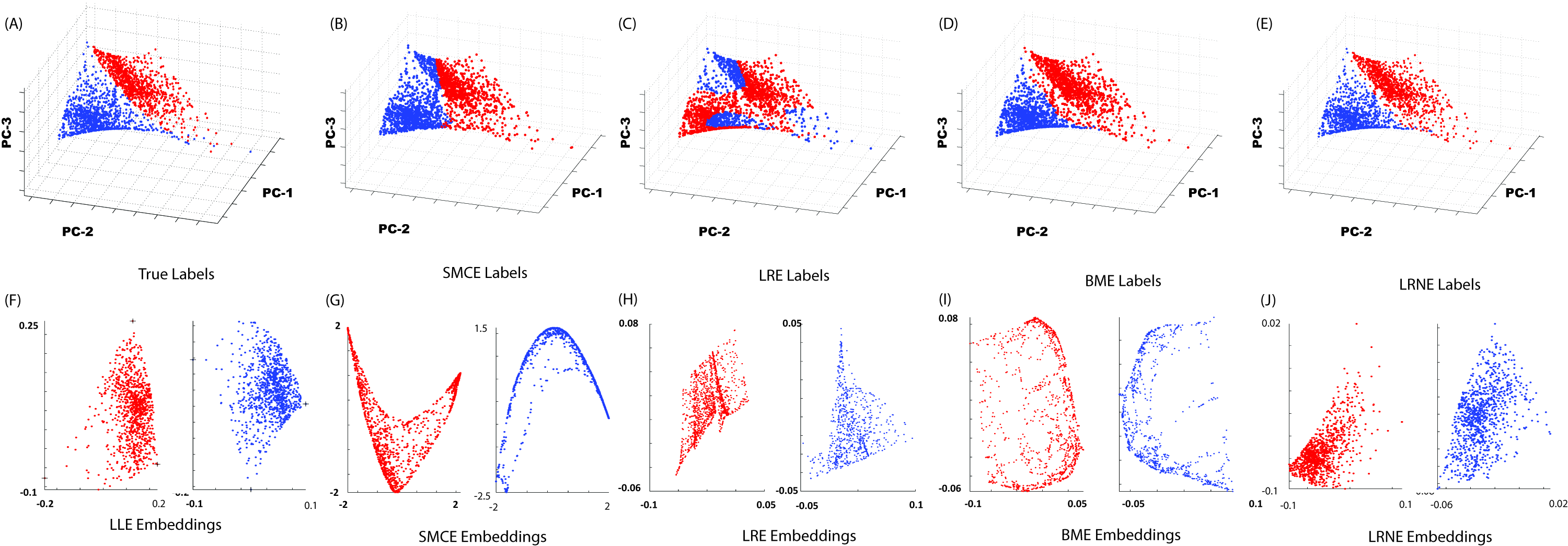}
		\caption{Simulated Hapke data-set (a) True Labels (b) SMCE class labels (c) LRE class labels (d) BME class labels (e) LRNE class labels (f) LLE embedding (g) SMCE embedding (h) LRE embedding (i) BME embedding (j) LRNE embedding}
		\label{fig_ternPerf}
	\end{figure*}

\begin{figure*}[!t]
		\centering
		\includegraphics[width=0.65\paperwidth]{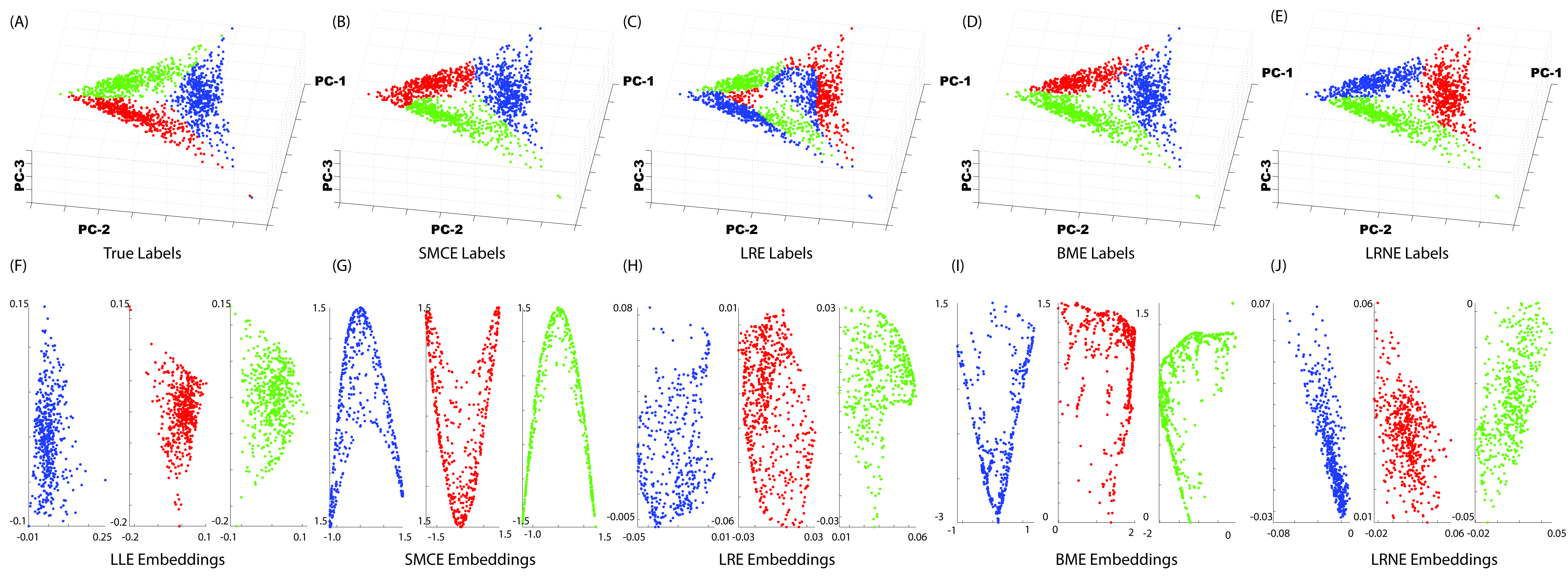}
		\caption{Three Simulated Hapke mixtures data-set (a) True Labels (b) SMCE class labels (c) LRE class labels (d) BME class labels (e) LRNE class labels (f) LLE embedding (g) SMCE embedding (h) LRE embedding (i) BME embedding (j) LRNE embedding}
		\label{fig_threeternPerf}
	\end{figure*}

	The data-set was chosen because of the high-dimensionality of the ambient space and the non-uniform sampling of the manifolds. The density gradient affects neighborhood structures at points with low density as even the nearest neighbors are quite far away making this a hard data-set for MCE algorithms. In particular this experiment focuses on two scenarios wherein there are (i) two ternary mixtures with four unique endmembers (i.e. two the endmembers are common to different ternary mixtures) and (ii) three ternary mixtures with four unique endmembers. For this experiment the spectra for the minerals olivine, ripidolite, illite and nontronite from the from the RELAB spectral database\footnote{RELAB Spectral Database: \textcopyright ~2008, Brown University, Providence, RI.; All Rights Reserved} were used as endmembers. In the first simulated experiment the dataset contains points lying on/near $2$ mixture-manifolds, with the boundary is composed of the mixed spectra between the shared endmembers. In this experiment the endmembers olivine and ripidolite are chosen to be common. $1003$ samples were generated by sampling each $2-D$ abundance simplex uniformly and mixed spectra was generated according to the Hapke Model. Fig. \ref{fig_ternPerf} (A) shows the first three PC's of the full dataset, points corresponding to the different mixtures are colored differently. The second experiment considers a simulated dataset with $3-$ternary mixtures formed with the same set of $4$ endmembers. In this scenario every pair of mixtures have $2$ endmembers in common. The different mixtures share different boundaries in common ($3$ in total) as shown in Fig. \ref{fig_threeternPerf} (A). For each ternary mixture $753$ samples were according to the Hapke Model in the same manner as described above.
	\subsubsection{Particulate Intimate Mixtures}
\begin{figure}[!t]
		\centering
		\includegraphics[width=0.8\columnwidth]{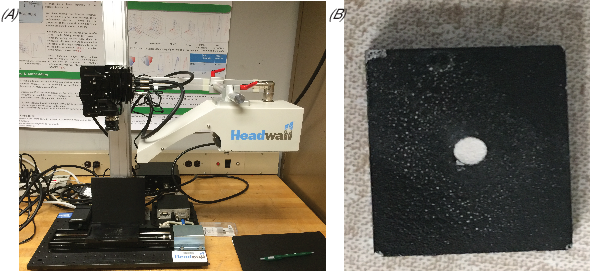}
		\caption{(A) The measurement setup for real Particulate Mixtures and (B) A prepared sample for measurement}
		\label{fig_measSetup}
	\end{figure}

\begin{figure*}[!t]
		\centering
		\includegraphics[width=0.5\paperwidth, height=2.0in]{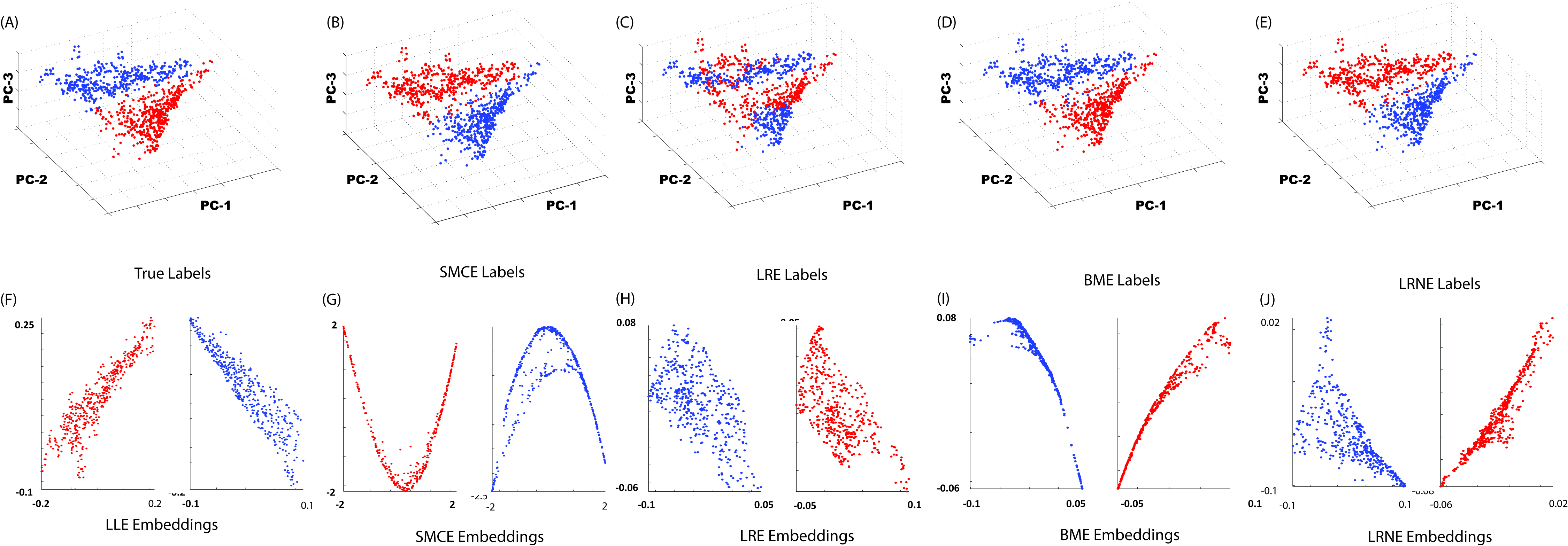}
		\caption{Two Real Particulate mixtures data-set (a) True Labels (b) SMCE class labels (c) LRE class labels (d) BME class labels (e) LRNE class labels (f) LLE embedding (g) SMCE embedding (h) LRE embedding (i) BME embedding (j) LRNE embedding}
		\label{fig_threeternRealPerf}
	\end{figure*}

	In addition to the simulated datasets, the MCE algorithms were also tested on real dataset with two ternary particulate mixtures with two endmembers in common. The minerals: olivine (San-Carlos), diopside and bytownite and augite were used as endmembers, olivine and diopside are the common endmembers. The augite mineral samples have a grain size of  $63-108~\mu m$, while the other mineral samples have a grain size range of $38-63~\mu m$. The (triangular) abundance simplices were then sampled $66$ times using a regular grid, intimate mixtures were prepared  corresponding to each abundance values. Each of these samples were then imaged using a Micro-Hyperspec\textsuperscript{\textregistered} SWIR M-Series imaging sensor\footnote{Micro-Hyperspec \textsuperscript{\textregistered} is a registered trademark of Headwall Photonics, Inc..} , to measure the sample reflectance in the wavelength range $(0.9-2.6~ \mu m)$ with a spectral resolution of $1.6~n m$. The setup for the measurement is shown in Fig. \ref{fig_measSetup} (A). Each prepared sample was placed in the Headwall Hyperspec Starter Kit (a prepared sample is shown in Fig. \ref{fig_measSetup} (B)). From each image some representative spectra corresponding to the sample were extracted and put into the dataset. The first three PC's of the two ternary mixtures dataset is shown in Fig. \ref{fig_threeternRealPerf} (A).
	\subsubsection{Hyperspectral Images of a Ceramic Tile Targets}
\begin{figure}[!t]
		\centering
		\includegraphics[width=0.65\columnwidth]{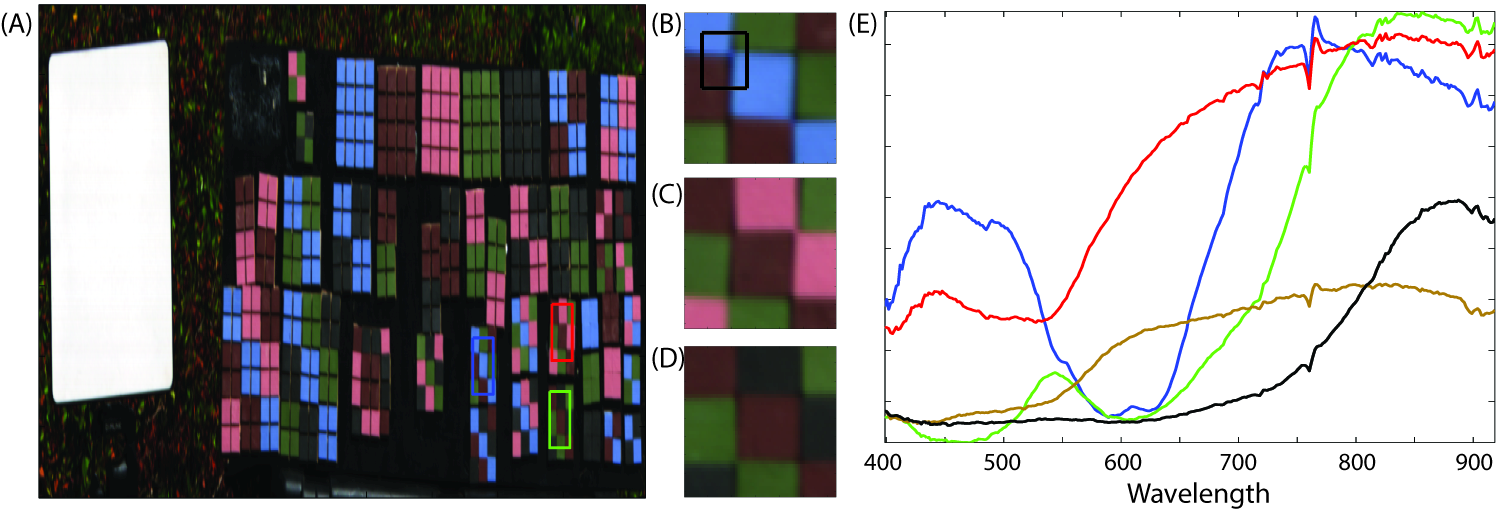}
		\caption{(A) RGB Composite of the Ceramic Tile Target (B) RGB closeups of the individual mixture targets (C) The endmember spectra from the target (The color of the plot matches the color of the tile)}
		\label{fig_VNIRTileTarget}
	\end{figure}

\begin{figure*}[!t]
		\centering
		\includegraphics[width=0.65\paperwidth]{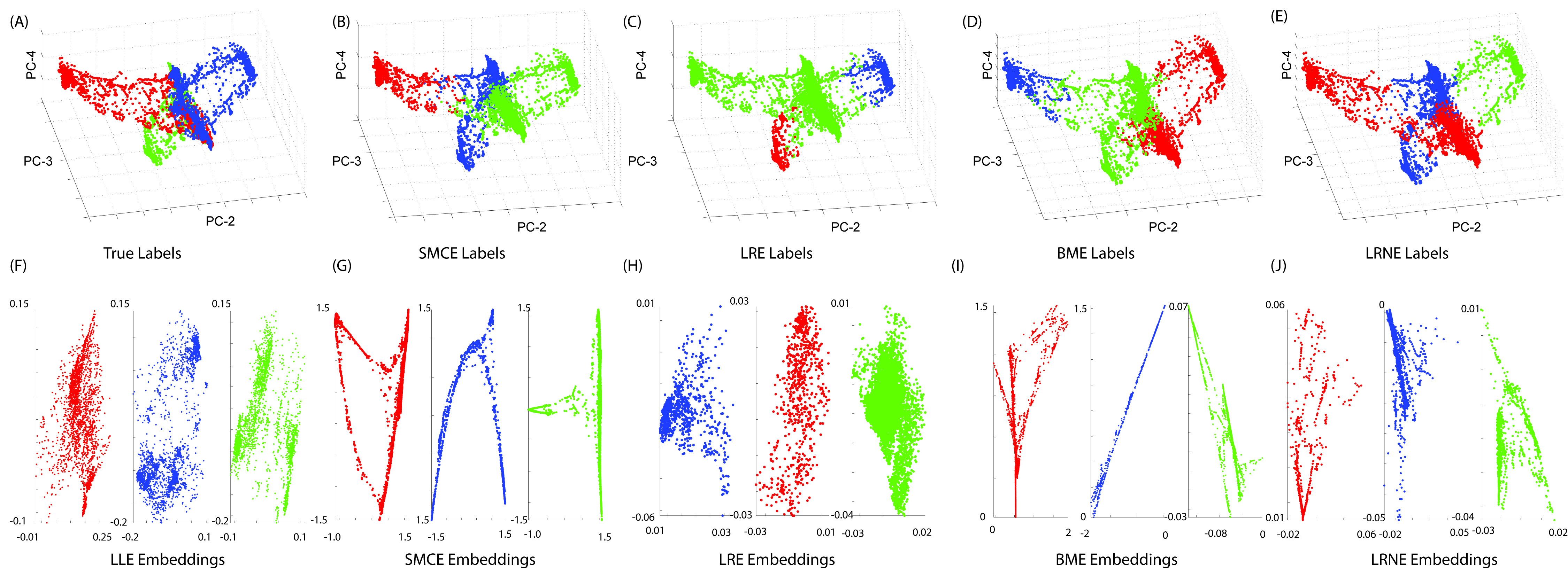}
		\caption{Two Real Particulate mixtures data-set (a) True Labels (b) SMCE class labels (c) LRE class labels (d) BME class labels (e) LRNE class labels (f) LLE embedding (g) SMCE embedding (h) LRE embedding (i) BME embedding (j) LRNE embedding}
		\label{fig_VNIRTileTarget_data}
	\end{figure*}

	The final dataset is a target of ceramic tiles. The target is shown in Fig. \ref{fig_VNIRTileTarget} (A). Each tile in the target is approximately $1.8~cm ~\times~ 1.8~cm$. The target was then placed outdoors and imaged using the Micro-Hyperspec\textsuperscript{\textregistered} VNIR E-Series imaging sensor\footnote{Micro-Hyperspec \textsuperscript{\textregistered} is a registered trademark of Headwall Photonics, Inc..} , to measure the sample reflectance in the wavelength range $(0.4-0.92~ \mu m)$ . The target was placed at a distance of $1.2~m$ which based on the properties of the imager leads to a pixel size of $0.18cm$. Experiments have shown that the Micro-Hyperspec\textsuperscript{\textregistered} VNIR E-Series imaging sensor has a points spread function of about $5 \times 7$ pixels-- which implies some linear mixing. Once the target was imaged radiance conversions were carried out and the approximate reflectance was calculated by ratioing each pixel's radiance-spectrum with the average radiance-spectrum of the Spectralon panel (the large white panel in Fig. \ref{fig_VNIRTileTarget} (A)) and multiplying each pixel-spectra with the known Spectralon reflectance spectra.  The end-member spectra for the various tiles is shown in Fig. \ref{fig_VNIRTileTarget} (E). From this image we choose three mixtures, \textit{Mixture-1} is made of the green, brown and blue tiles (highlighted with the blue box in Fig. \ref{fig_VNIRTileTarget} (A)), \textit{Mixture-2} is made of the green, brown and blue tiles (highlighted with the red box in Fig. \ref{fig_VNIRTileTarget} (A)) and \textit{Mixture-3} is made of the green, brown and black tiles (highlighted with the green box in Fig. \ref{fig_VNIRTileTarget} (A)). A closeup view of the mixtures is shown in Fig. \ref{fig_VNIRTileTarget} (B)-(D)).
	The sub-image corresponding to each picture is $41 \times 63$ which means that there are $2583$ samples corresponding to each mixture. The first three PC's of this data is shown in Fig. \ref{fig_VNIRTileTarget_data} (A). This dataset is the most complex dataset considered in these experiments. Firstly, the dataset is measured outdoors, which has some distortion effects such as atmosphere, variable lighting conditions etc.. Secondly due to the regular shape of the target, at the intersection between multiple tiles as shown in Fig. \ref{fig_VNIRTileTarget} (B)), there is always more of one end-member as compared to the others, due to this most of the pixels are quite close to the facets and there is a large hole at the center of the abundance simplex which can also be Fig. \ref{fig_VNIRTileTarget_data} (A). Thirdly, the tile is larger than the estimated point spread function, this causes there to be a large number of pure and binary mixed pixels leading to very high density of points at the facets of the simplex. Fourth, the tile has variable roughness due to this there is significant variation in even the endmember spectra. Due to these various factors this dataset suffers from more distortions the other datasets.
	\subsection{Experiments Setup}
	The various algorithms are then tested on the different datasets over variety of different settings. All the algorithms under test feature a parameter that trades off the reconstruction objective with a penalty on using points on the ``wrong" manifold. The BME, SMCE and LRNE are graph-based and require a parameter $k$, the number of neighbors in the $k$-NN graph. The BME also requires additional information on the dimensionality of the manifold each point belongs to. For LRE, SMCE and LRNE we try the algorithms for a variety of values of $\lambda$, in particular we tried the values $[0.001~0.005~0.01~0.05~0.1~0.5~1~5~10]$. For the BME we set the trade-off parameter $\gamma$ from the set $[0.75~0.85~0.9~0.925~0.95~0.99~0.995~1]$. The neighborhood size is varied between $25$ and $50$ in steps of $5$.
	\subsection{Clustering Performance}
	The effect of the parameters on algorithm performance is shown in Table. \ref{tab:ParamsPerf_class} for the first simulated dataset. The LRE shows near constant performance and does not change significantly with the change in the parameters. The BME and the SMCE on the other hand are more sensitive to the parameter $\lambda$/$\gamma$. Over a wide range of parameters the LRNE show significant improvement over its competitors across the different datasets. Across the different datasets the effect of the parameters are quite similar. The best classification performance over the range of parameter values each algorithm for the different datasets is shown in Table. \ref{tab:BestClassPerf}. 
	
	For the \textit{simulated dataset with $2$ ternary mixtures} the classification performance of the LRNE (shown in Fig. \ref{fig_ternPerf} (E)) outperforms its competitors. The LRE fails in terms of classification (as shown in Fig. \ref{fig_ternPerf} (C)), while the SMCE and BME have some success in identifying the different mixtures it is less stable in the regions near the boundary with significant density variation (shown in Fig. \ref{fig_ternPerf} (B) \& (D) respectively). A similar trend can be seen in the case of \textit{simulated dataset with $3$ ternary mixtures} and the \textit{Particulate dataset with $2$ ternary mixtures} as shown in Figs. \ref{fig_threeternRealPerf} \& \ref{fig_threeternRealPerf} (B-E). In the case of the \textit{Ceramic Tile Target} dataset none of the algorithms are very successful and all have misclassification rates close to ~40\%. While the numbers are underwhelming it quite encouraging to note the LRNE in successful in identifying the various mixtures in the data (shown in Fig. \ref{fig_VNIRTileTarget_data} (E)). In fact the only misclassification seem occurs in the intersection. On the other hand the SMCE and BME while somewhat successful in identifying the arms (shown in Fig. \ref{fig_VNIRTileTarget_data} (B) \& (D)) still mis-classifies spectra farther into the arms as compared to the LRNE. While the points at the intersection (i.e. spectra corresponding to mixtures of the Green and Brown tile) are considered misclassified in this experiment it is important to note that these binary combination which lie on the intersection of the different mixture-manifolds are a part of all three mixture-manifolds. Thus these ``errors" are not as egregious as placing points deeper into the arms in wrong classes.
	
	The LRE \cite{LRE} clustering fails over all the datasets, based on this we can deduce that LRE scheme of using all points in reconstruction and the assumption of similarity in reconstruction coefficient structure for points on the same manifold does not hold. The sparsity assumption of SMCE \cite{SMCE} does reasonably well in terms of clustering, but in the presence of variable density when the nearest neighbors are not from the same manifold the clustering performance falls off a little. Inspite of having additional information on the dimensionality of the different manifolds the BME does not always improve on the performance of the other algorithms. Using local techniques and variable neighborhood size the LRNE is better equipped to deal with both non-linearities and variable density.
	\begin{table*}[]
		\centering
		\caption{Effect of parameters on Classification performance of the simulated $2$ mixture dataset}
		\label{tab:ParamsPerf_class}
		\begin{tabular}{|l|l|l|l|l|l|l|l|l|l|l|}
			\hline
			&  & \begin{tabular}[c]{@{}l@{}}$\lambda=0.001$\\ $\gamma=0.75$\end{tabular} & \begin{tabular}[c]{@{}l@{}}$\lambda=0.005$\\ $\gamma=0.80$\end{tabular} & \begin{tabular}[c]{@{}l@{}}$\lambda=0.01$\\ $\gamma=0.85$\end{tabular} & \begin{tabular}[c]{@{}l@{}}$\lambda=0.05$\\ $\gamma=0.925$\end{tabular} & \begin{tabular}[c]{@{}l@{}}$\lambda=0.1$\\ $\gamma=0.925$\end{tabular} & \begin{tabular}[c]{@{}l@{}}$\lambda=0.5$\\ $\gamma=0.95$\end{tabular} & \begin{tabular}[c]{@{}l@{}}$\lambda=1$\\ $\gamma=0.99$\end{tabular} & \begin{tabular}[c]{@{}l@{}}$\lambda=5$\\ $\gamma=0.995$\end{tabular} & \begin{tabular}[c]{@{}l@{}}$\lambda=10$\\ $\gamma=1$\end{tabular} \\ \hline
			\multirow{3}{*}{k=30} & BME & 44.87 & 45.31 & 39.48 & 39.93 & 36.59 & 35.99 & 32.15 & 28.61 & 49.90 \\ \cline{2-11} 
			& SMCE & 35.54 & 34.60 & 32.60 & 26.67 & 25.07 & 21.49 & 24.19 & 22.13 & 21.49 \\ \cline{2-11} 
			& LRNE & 49.90 & 49.90 & 2.84 & 10.17 & 12.86 & 19.19 & 24.18 & 28.36 & 28.56 \\ \hline
			\multirow{3}{*}{k=35} & BME & 43.17 & 43.12 & 36.19 & 36.24 & 36.74 & 37.89 & 35.24 & 35.24 & 27.02 \\ \cline{2-11} 
			& SMCE & 37.24 & 36.04 & 34.65 & 26.87 & 25.02 & 21.29 & 23.33 & 22.98 & 21.24 \\ \cline{2-11} 
			& LRNE & 49.90 & 49.90 & 2.49 & 8.47 & 11.47 & 17.70 & 22.68 & 27.72 & 27.97 \\ \hline
			\multirow{3}{*}{k=40} & BME & 45.1 & 43.82 & 43.72 & 43.82 & 44.62 & 44.72 & 34.85 & 35.29 & 27.12 \\ \cline{2-11} 
			& SMCE & 38.19 & 37.74 & 36.49 & 27.57 & 26.27 & 21.88 & 22.28 & 22.98 & 21.24 \\ \cline{2-11} 
			& LRNE & 49.90 & 49.90 & 2.54 & 7.28 & 10.77 & 17.35 & 22.28 & 27.77 & 28.51 \\ \hline
			\multirow{3}{*}{k=45} & BME & 46.46 & 45.21 & 45.11 & 45.76 & 46.26 & 33.90 & 31.21 & 29.76 & 49.90 \\ \cline{2-11} 
			& SMCE & 39.03 & 38.68 & 37.74 & 29.61 & 27.37 & 22.48 & 23.13 & 24.68 & 23.13 \\ \cline{2-11} 
			& LRNE & 49.90 & 49.90 & 2.64 & 5.68 & 9.82 & 16.30 & 22.28 & 28.86 & 29.71 \\ \hline
			\multirow{3}{*}{k=50} & BME & 45.66 & 45.46 & 43.67 & 43.72 & 44.22 & 44.52 & 36.04 & 36.04 & 26.57 \\ \cline{2-11} 
			& SMCE & 40.13 & 39.78 & 39.18 & 30.96 & 29.06 & 25.97 & 25.42 & 25.67 & 23.33 \\ \cline{2-11} 
			& LRNE & 49.90 & 49.90 & 2.69 & 4.54 & 8.18 & 15.25 & 21.34 & 29.16 & 29.81 \\ \hline
			& LRE & 46.96 & 46.96 & 46.96 & 46.96 & 46.93 & 44.28 & 42.81 & 44.01 & 44.72 \\ \hline
		\end{tabular}
	\end{table*}
	
	\begin{table}[]
		\centering
		\caption{Best Classification Performance across the various datasets}
		\label{tab:BestClassPerf}
		\begin{tabular}{|l|l|l|l|l|}
			\hline
			Dataset                       & LRE   & BME   & SMCE  & LRNE  \\ \hline
			Simulated Dataset - 2 Ternary & 42.81 & 5.13 & 14.56 & 2.49 \\ \hline
			Simulated Dataset - 3Ternary  & 41.25 & 7.09  & 11.13 & 6.23  \\ \hline
			Real Particulate Mixture      & 39.96 & 1.19 & 10.82  & 0.97  \\ \hline
			Hyperspectral Tile Target     & 46.68 & 41.72 & 42.12 & 40.86 \\ \hline
		\end{tabular}
	\end{table}
	\subsection{Embedding Performance}
	The embeddings generated by the various algorithms for the different datasets is shown in Figs. \ref{fig_ternPerf}, \ref{fig_threeternPerf}, \ref{fig_threeternRealPerf} and \ref{fig_VNIRTileTarget_data} (G)-(J). Since all the different different datasets contain only ternary mixtures, we expect the embedding of each mixture-manifold to be a $2-D$ simplex, i.e. a triangle. The embeddings generated by LRE are untrustworthy due to its poor classification performance. In the case of the BME the embeddings are highly distorted (when classification performance is reasonable) as the reconstruction matrix which is a thresholded reconstruction matrix does not capture the geometric information accurately. The SMCE embeddings show significant distortions. The embedding technique used by all these algorithms is local, i.e. it only preserves the geometric information captured in each neighborhood, and there is no penalty paid for distorting the relative position or shape of one neighborhood with respect to another if they have no overlap. The SMCE creates very sparse neighborhoods and there is very little overlap between neighborhoods, which means there is a very small penalty for distorting the relative shapes and position of different neighborhoods, leading to embeddings with highly distorted global shapes. The embeddings generated by the LRNE are far more reasonable and bear closer resemblance to the expected triangular shape. 
	
	In addition, we also attempted to make a quantitative comparison of the embedding performance of the LRNE to some ``optimal" embeddings. Since most of the reconstruction matrix-based schemes are based on the LLE, the LLE embeddings were considered optimal, and the embeddings generated by the LRNE were compared to LLE embedding of each manifold separately. In the best scenario, the embeddings generated by the ``optimal" LLE will be an approximate representation of the parameter space ``up to some affine transformation" \cite{goldberg2008manifold}. To check this we will estimate the abundances in the emedded space using Fully constrained least squares \cite{heinz2001fully}, i.e from the embeddings we find the estimated weights $W_{i}^{\mathrm{est}}$ for the embedded points $y_i$ with respect to the embedded versions of the endmembers $V_y$. We define the average error in the embedding of a manifold as $\frac{1}{3n}\sum_{i=1}^{n} \norm{W^{\mathrm{est}}_i - W_i}$, where $W_i$ are the true abundances \& $n$ is the number correctly classified points in each manifold. [Note: - we do not consider incorrectly classified points as the embedding error in these cases the error is affected by the classification]. The effect of the parameters $\lambda$ and the number of neighbors $k$ on the embedding error of the LRNE, is shown for one of the mixture-manifolds in the \textit{tow simulated ternary mixtures} dataset in Table. \ref{tab:ParamsEffect_Embedding}.
	\begin{table*}[]
		\centering
		\caption{Effect of parameters on LRNE embedding of manifold-$1$}
		\label{tab:ParamsEffect_Embedding}
		\begin{tabular}{|l|l|l|l|l|l|l|l|}
			\hline
			& $\lambda=0.01$ & $\lambda=0.05$ & $\lambda=0.1$ & $\lambda=0.5$ & $\lambda=1$ & $\lambda=5$ & LLE \\ \hline
			k=30 & 0.088 & 0.1028 & 0.0946 & 0.1383 & 0.1738 & 0.0801 & 0.0393 \\ \hline
			k=35 & 0.1068 & 0.1028 & 0.1012 & 0.0889 & 0.1830 & 0.0613 & 0.0389 \\ \hline
			k=40 & 0.0952 & 0.1014 & 0.1062 & 0.0821 & 0.1734 & 0.0519 & 0.0388 \\ \hline
			k=45 & 0.0957 & 0.0967 & 0.1001 & 0.0792 & 0.1610 & 0.0447 & 0.0388 \\ \hline
			k=50 & 0.0984 & 0.0948 & 0.0866 & 0.0887 & 0.1507 & 0.0399 & 0.0389 \\ \hline
		\end{tabular}
	\end{table*}
	
	\begin{table}[]
		\centering
		\caption{Avg. Error in embedding performance}
		\label{tab:embErr}
		\begin{tabular}{|l|l|l|l|}
			\hline
			&  & LLE & LRNE \\ \hline
			\multirow{2}{*}{\begin{tabular}[c]{@{}l@{}}Simulated 2 \\ Ternary Mixture\end{tabular}} & mixture-1 & 0.0388 & 0.0399 \\ \cline{2-4} 
			& mixture-2 & 0.443 & 0.0525 \\ \hline
			\multirow{3}{*}{\begin{tabular}[c]{@{}l@{}}Simulated 3\\ Ternary Mixtures\end{tabular}} & mixture-1 &  &  \\ \cline{2-4} 
			& mixture-2 &  &  \\ \cline{2-4} 
			& mixture-3 &  &  \\ \hline
			\multirow{2}{*}{\begin{tabular}[c]{@{}l@{}}Particulate 2\\ Ternary Mixtures\end{tabular}} & mixture-1 & 0.0962 & 0.115 \\ \cline{2-4} 
			& mixture-2 & 0.0688 & 0.0687 \\ \hline
		\end{tabular}
	\end{table}
	In general we note that the embedding error is slightly higher than the corresponding LLE error. For very small values of $\lambda$ incorrect points are given high priority in the reconstruction this leading to distortion. At larger values of $\lambda$ there is better classification but higher reconstruction error. The best embedding error for the different simulated manifolds is shown in Table \ref{tab:embErr}. In general the best embedding error from the LRNE is very close to the individual LLE embedding errors for each manifold as shown in Table. \ref{tab:embErr}.
	
	\section{Conclusion} 
	\label{conclusion}
	The Low Rank Neighborhood algorithm successfully generates a reconstruction matrix that can be used for both manifold clustering and embedding for hyperspectral mixture manifolds. The LRNE outperforms existing state-of-the-art MCE algorithms over a variety of simulated and real data-sets. The LRNE shows improved clustering especially in scenarios where there are local variation in densities. Additionally since the LRNE allows the users to choose the size of the neighborhood $k$, we can ensure that there is enough overlap between different neighborhood patches which ensures that the LRNE is better able to retrieve the global shape of the abundance space. The embeddings generated by the LRNE compare favorably to the ones generated by dedicated embedding algorithms on each mixture manifold separately.
	\appendices
	\section{Matrix Norm Propositions}
	\textit{Lemma 1:} The norm propositions for the different norms are: 
	\begin{multline}
	~~~~~~~~~~~~~~~~~~\|z\|_{\infty} \leq \|z\|_{1} \leq n~\|z\|_{\infty} \\
	~~~~\|z\|_{\infty} \leq \|z\|_{2} \leq \sqrt{n}~\|z\|_{\infty} \\
	~\|z\|_{2} \leq \|z\|_{1} \leq \sqrt{n}~\|z\|_{2} \\
	~~~~\|z\|_{2} \leq \|z\|_{\Omega_{D}} \leq \sqrt{n}~\|z\|_{2}\\
	~~~~~~~\|z\|_{\Omega_{D}} \leq \|z\|_{1} \leq \sqrt{n}~\|z\|_{\Omega_{D}}\\
	\label{eqn:NormProp}
	\end{multline}
	\textit{Proof:} The first three are well known propositions as has been in mentioned in \cite{peng2016constructing}. The other two propositions are simple extensions of the  of \textbf{\textit{Proposition-3}} in \cite{grave2011trace} which is:\\
	\begin{equation*}
	\|w\|_{2} \leq \|w\|_{\Omega_{D}} \leq \|w\|_{1}
	\end{equation*}
	\section{Proof of Lemma 2}
	\label{sec:ProofLemma2}
	\textit{Lemma 2:} For a point $y$ defined as in Lemma 1, we will have that $\|z_{D_{x}}\|_p~<~\|z_{D_{-x}}\|_p$ if
	\begin{equation}
	\sigma_{min}(D_{x})~ \geq~ k_{p}~ cos(\theta_{min})~ \|D_{x}\|_{max,2} 
	\end{equation}
	where $\sigma_{min}(D_{x})$ is the smallest singular value of $D_{x}$, $\theta_{min}$ is the first principal angle between $D_{x}$ and $D_{-x}$, and $\|D_{x}\|_{max,2}$ is the maximum $\ell_{2}$ of the columns of $D_{-x}$.\\
	\textit{Proof:} Since $y \in \{\mathcal{S}|\mathcal{S} = \mathcal{S}_{D_{x}} \cap \mathcal{S}_{D_{-x}} \}$, we can write $y = D_{x} z_{D_{x}}$, using the SVD we have that $D_{x} = U_{r_{x}} \Sigma_{r_{x}} V{^{T}_{r_{x}}}$, where $r_{x}$ is the rank of the matrix $D_{x}$ (and the number of non-zero entries in the matrix $\Sigma_{r_{x}}$), i.e. $\Sigma_{r_{x}} = diag(\sigma_1(D_{x}),~\sigma_2(D_{x}),.~.~.~.,~\sigma_{r_{x}}(D_{x}))$. Now, based on this we can write:
	\begin{multline*}
	~~~~~~~~~~~~~~~~~~~~~y = U_{r_{x}} \Sigma_{r_{x}} V{^{T}_{r_{x}}} z_{D_{x}}\\
	z_{D_{x}} =  V{_{r_{x}}} \Sigma{_{r_{x}}^{-1}} U{^{T}_{r_{x}}} y~~~~~~~~~~~~~~~~~~~~~~
	\end{multline*}
	Based on the proposition in Eqn. (\ref{eqn:NormProp}) we have
	\begin{equation}
	\|z_{D_{x}}\|_{p} ~ \leq ~ \|z_{D_{x}}\|_{1} ~ \leq ~ \sqrt{n} \|z_{D_{x}}\|_{2} = \sqrt{n} \|V_{r_{x}} \Sigma{_{r_{x}}^{-1}} U{_{r_{x}}^{T}} y\|_{2}
	\end{equation}
	Since $\frac{n}{r_{x}} \geq 1$ and the Frobenius norm is subordinate to the Euclidean vector norm:
	\begin{multline}
	\|z_{D_{x}}\|_{p} ~ \leq ~  \sqrt{n} \|V_{r_{x}} \Sigma{_{r_{x}}^{-1}} U{_{r_{x}}^{T}}\|_{F} \|y\|_{2}\\
	~~~~~~~~ \leq \dfrac{\sqrt{n}}{\sqrt{\sigma{^2_{1}}(D_{x})+.~.~.~+\sigma{^2_{r_{x}}}(D_{x})}} \|x\|_{2} \leq \sigma{^{-1}_{min}}(D_{x}) \|y\|_{2} 
	\label{eqn:LHS}
	\end{multline}
	where $\sigma{_{min}}(D_{x})$ is the smallest singular value of the matrix $D_x$.\\
	Also , since $y \in \{\mathcal{S}|\mathcal{S} = \mathcal{S}_{D_{x}} \cap \mathcal{S}_{D_{-x}} \}$, we can also represent $y$ as a linear combination of $D_{-x}$, since it lies on the intersection of the subspaces, i.e., we have that
	\begin{equation*}
	y = D_{-x} z_{D_{-x}}
	\end{equation*}
	\begin{equation}
	\|y\|{^{2}_{2}} ~= ~y^Ty = y^T D_{-x} z_{D_{-x}}
	\label{eqn:l2_vecNorm}
	\end{equation}
	Now using H\"{o}lder's inequality we can say that:
	\begin{equation}
	\| y^T D_{-x} z_{D_{-x}} \|{_{1}} \leq \|y^T D_{-x} \|{_{\infty}} \|z_{D_{-x}}\|{_{1}} \leq \|D{^{T}_{-x}} y \|{_{\infty}} \|z_{D_{-x}}\|{_{1}}
	\end{equation}
	now substituting this result in Eqn. (\ref{eqn:l2_vecNorm})
	\begin{equation}
	\|y\|{_{2}^{2}} \leq \|D{^{T}_{-x}} x \|{_{\infty}} \|z_{D_{-x}}\|{_{1}}
	\label{eqn:l2_vecNorm_1}
	\end{equation}
	now we can define $\|D{^{T}_{-x}} x \|{_{\infty}}$ as:
	\begin{multline}
	~~~\|D{^{T}_{-x}} y \|{_{\infty}} = max \lbrace \vert [D_{-x}]{_{1}^{T}} y \vert,~ \vert [D_{-x}]{_{2}^{T}} y \vert,~.~.~.~. \rbrace \\	
	\leq \|D_{-x}\|_{max,2}\|y\|_{2} ~~cos~ \theta_{min} \\
	\end{multline}
	where, $[D_{-x}]{_{i}}$, is the $i^{th}$ column of the matrix $D_{-x}$, $\theta_{min}$ is the first principal angle between $\mathcal{S}_{D_{x}}$ and $\mathcal{S}_{D_{-x}}$ and $\|D_{-x}\|_{max,2}$ denotes the maximum $\ell_{2}-$norm of the columns of $D_{-x}$. Substituting this result in Eqn. (\ref{eqn:l2_vecNorm_1}) we have that:
	\begin{equation*}
	\|y\|{_{2}^{2}} \leq \|D_{-x}\|_{max,2}~\|y\|_{2} ~cos( \theta_{min}) \|z_{D_{-x}}\|_{1}
	\end{equation*}
	\begin{equation*}
	\|y\|{_{2}} \leq \|D_{-x}\|_{max,2} ~~cos (\theta_{min})~\|z_{D_{-x}}\|_{1}
	\end{equation*}
	hence
	\begin{equation}
	\|z_{D_{-x}}\|_{1} \geq \dfrac{	\|y\|{_{2}}}{\|D_{-x}\|_{max,2} ~~cos (\theta_{min})}
	\end{equation}
	from the proposition on the $p-norms$, we have
	\begin{equation*}
	k_p~\|z_{D_{-x}}\|_{p} \geq \dfrac{	\|y\|{_{2}}}{\|D_{-x}\|_{max,2} ~~cos (\theta_{min})}
	\end{equation*}
	where $k_p = n$ when $p=\infty$ and $k_p = \sqrt{n}$ otherwise.
	\begin{equation}
	\|z_{D_{-x}}\|_{p} \geq \dfrac{	\|x\|{_{2}}}{k_p~\|D_{-x}\|_{max,2} ~~cos (\theta_{min})}
	\end{equation}
	For the condition $\|z_{D_{x}}\|_{p} < \|z_{D_{-x}}\|_{p}$, then
	\begin{equation*}
	\sigma{^{-1}_{min}}(D_x) \|y\|_{2} < \dfrac{	\|y\|{_{2}}}{n~\|D_{-x}\|_{max,2} ~~cos (\theta_{min})}
	\end{equation*}
	then
	\begin{equation}
	\sigma{_{min}}(D_x)  > {k_p~\|D_{-x}\|_{max,2} ~cos (\theta_{min})}
	\label{eqn:FinalIPD}
	\end{equation}


	
	%



	\ifCLASSOPTIONcaptionsoff
	\newpage
	\fi

	
	
	\bibliographystyle{plain}
	\bibliography{example_paper}

\begin{thebibliography}{10}

\bibitem{Belkin01laplacianeigenmaps}
Mikhail Belkin and Partha Niyogi.
\newblock Laplacian eigenmaps and spectral techniques for embedding and
  clustering.
\newblock In {\em Advances in Neural Information Processing Systems 14}, pages
  585--591. MIT Press, 2001.

\bibitem{unmixingLinearReview}
Jos{\'e}~M Bioucas-Dias, Antonio Plaza, Nicolas Dobigeon, Mario Parente, Qian
  Du, Paul Gader, and Jocelyn Chanussot.
\newblock Hyperspectral unmixing overview: Geometrical, statistical, and sparse
  regression-based approaches.
\newblock {\em Selected Topics in Applied Earth Observations and Remote
  Sensing, IEEE Journal of}, 5(2):354--379, 2012.

\bibitem{ADMM}
Stephen Boyd, Neal Parikh, Eric Chu, Borja Peleato, and Jonathan Eckstein.
\newblock Distributed optimization and statistical learning via the alternating
  direction method of multipliers.
\newblock {\em Foundations and Trends{\textregistered} in Machine Learning},
  3(1):1--122, 2011.

\bibitem{cai2010singular}
Jian-Feng Cai, Emmanuel~J Cand{\`e}s, and Zuowei Shen.
\newblock A singular value thresholding algorithm for matrix completion.
\newblock {\em SIAM Journal on Optimization}, 20(4):1956--1982, 2010.

\bibitem{Candes:2009:EMC:1666896.1666899}
Emmanuel~J. Cand\'{e}s and Benjamin Recht.
\newblock Exact matrix completion via convex optimization.
\newblock {\em Found. Comput. Math.}, 9(6):717--772, December 2009.

\bibitem{chang2007hyperspectral}
Chein-I Chang.
\newblock {\em Hyperspectral data exploitation: theory and applications}.
\newblock John Wiley \& Sons, 2007.

\bibitem{eismann2012hyperspectral}
Michael~Theodore Eismann.
\newblock Hyperspectral remote sensing.
\newblock SPIE Bellingham, 2012.

\bibitem{NIPS2011_4246}
Ehsan Elhamifar and Ren\'{e} Vidal.
\newblock Sparse manifold clustering and embedding.
\newblock In J.~Shawe-Taylor, R.S. Zemel, P.L. Bartlett, F.~Pereira, and K.Q.
  Weinberger, editors, {\em Advances in Neural Information Processing Systems
  24}, pages 55--63. Curran Associates, Inc., 2011.

\bibitem{elhamifar2013sparse}
Ehsan Elhamifar and Rene Vidal.
\newblock Sparse subspace clustering: Algorithm, theory, and applications.
\newblock {\em IEEE transactions on pattern analysis and machine intelligence},
  35(11):2765--2781, 2013.

\bibitem{goldberg2008manifold}
Yair Goldberg, Alon Zakai, Dan Kushnir, and Ya’acov Ritov.
\newblock Manifold learning: The price of normalization.
\newblock {\em Journal of Machine Learning Research}, 9(Aug):1909--1939, 2008.

\bibitem{golub2012matrix}
Gene~H Golub and Charles~F Van~Loan.
\newblock {\em Matrix computations}, volume~3.
\newblock JHU Press, 2012.

\bibitem{gong2012robust}
Dian Gong, Xuemei Zhao, and G{\'e}rard Medioni.
\newblock Robust multiple manifolds structure learning.
\newblock {\em arXiv preprint arXiv:1206.4624}, 2012.

\bibitem{grave2011trace}
Edouard Grave, Guillaume~R Obozinski, and Francis~R Bach.
\newblock Trace lasso: a trace norm regularization for correlated designs.
\newblock In {\em Advances in Neural Information Processing Systems}, pages
  2187--2195, 2011.

\bibitem{green1998imaging}
Robert~O Green, Michael~L Eastwood, Charles~M Sarture, Thomas~G Chrien, Mikael
  Aronsson, Bruce~J Chippendale, Jessica~A Faust, Betina~E Pavri, Christopher~J
  Chovit, Manuel Solis, et~al.
\newblock Imaging spectroscopy and the airborne visible/infrared imaging
  spectrometer (aviris).
\newblock {\em Remote Sensing of Environment}, 65(3):227--248, 1998.

\bibitem{hapkeBook}
Bruce Hapke.
\newblock {\em Theory of Reflectance and Emittance Spectroscopy}.
\newblock Cambridge University Press, second edition, 2012.
\newblock Cambridge Books Online.

\bibitem{heinz2001fully}
Daniel~C Heinz et~al.
\newblock Fully constrained least squares linear spectral mixture analysis
  method for material quantification in hyperspectral imagery.
\newblock {\em IEEE transactions on geoscience and remote sensing},
  39(3):529--545, 2001.

\bibitem{huo2007survey}
Xiaoming Huo, Xuelei~Sherry Ni, and Andrew~K Smith.
\newblock A survey of manifold-based learning methods.
\newblock {\em Recent advances in data mining of enterprise data}, pages
  691--745, 2007.

\bibitem{keshava2002spectral}
Nirmal Keshava and John~F Mustard.
\newblock Spectral unmixing.
\newblock {\em IEEE signal processing magazine}, 19(1):44--57, 2002.

\bibitem{li2010learning}
Chun-guang Li, Jun Guo, and Hong-gang Zhang.
\newblock Learning bundle manifold by double neighborhood graphs.
\newblock {\em Computer Vision--ACCV 2009}, pages 321--330, 2010.

\bibitem{liangrocapart1998mixed}
S~Liangrocapart and Maria Petrou.
\newblock Mixed pixels classification.
\newblock In {\em Remote Sensing}, pages 72--83. International Society for
  Optics and Photonics, 1998.

\bibitem{lin2010augmented}
Zhouchen Lin, Minming Chen, and Yi~Ma.
\newblock The augmented lagrange multiplier method for exact recovery of
  corrupted low-rank matrices.
\newblock {\em arXiv preprint arXiv:1009.5055}, 2010.

\bibitem{LADMAP}
Zhouchen Lin, Risheng Liu, and Zhixun Su.
\newblock Linearized alternating direction method with adaptive penalty for
  low-rank representation.
\newblock In {\em Advances in neural information processing systems}, pages
  612--620, 2011.

\bibitem{LRR}
Guangcan Liu, Zhouchen Lin, and Yong Yu.
\newblock Robust subspace segmentation by low-rank representation.

\bibitem{Liu:2011}
Risheng Liu, Ruru Hao, and Zhixun Su.
\newblock Mixture of manifolds clustering via low rank embedding.
\newblock {\em Journal of Information and Computational Science},
  8(5):725--737, 2011.

\bibitem{lu2013correlation}
Canyi Lu, Jiashi Feng, Zhouchen Lin, and Shuicheng Yan.
\newblock Correlation adaptive subspace segmentation by trace lasso.
\newblock In {\em Proceedings of the IEEE International Conference on Computer
  Vision}, pages 1345--1352, 2013.

\bibitem{kMeans}
James MacQueen et~al.
\newblock Some methods for classification and analysis of multivariate
  observations.
\newblock In {\em Proceedings of the fifth Berkeley symposium on mathematical
  statistics and probability}, volume~1, pages 281--297. Oakland, CA, USA.,
  1967.

\bibitem{murchie2007compact}
Scott Murchie, R~Arvidson, Peter Bedini, K~Beisser, J-P Bibring, J~Bishop,
  J~Boldt, P~Cavender, T~Choo, RT~Clancy, et~al.
\newblock Compact reconnaissance imaging spectrometer for mars (crism) on mars
  reconnaissance orbiter (mro).
\newblock {\em Journal of Geophysical Research: Planets}, 112(E5), 2007.

\bibitem{peng2016constructing}
Xi~Peng, Zhiding Yu, Zhang Yi, and Huajin Tang.
\newblock Constructing the l2-graph for robust subspace learning and subspace
  clustering.
\newblock {\em IEEE transactions on cybernetics}, 2016.

\bibitem{pieters2009moon}
Carle~M Pieters, Joseph Boardman, Bonnie Buratti, Alok Chatterjee, Roger Clark,
  Tom Glavich, Robert Green, James Head~III, Peter Isaacson, Erick Malaret,
  et~al.
\newblock The moon mineralogy mapper (m$^3$) on chandrayaan-1.
\newblock {\em Current Science}, pages 500--505, 2009.

\bibitem{LLE}
Sam~T. Roweis and Lawrence~K. Saul.
\newblock Nonlinear dimensionality reduction by locally linear embedding.
\newblock {\em Science}, 290:2323--2326, 2000.

\bibitem{JMCE}
A.~M. Saranathan and M.~Parente.
\newblock Simultaneous clustering and embedding for multiple intimate mixtures.
\newblock In {\em Geoscience and Remote Sensing Symposium (IGARSS), 2015 IEEE
  International}, pages 2397--2400, July 2015.

\bibitem{Saranathan14ManifoldClustering}
Arun Saranathan and Mario Parente.
\newblock Manifold clustering based unmixing for the multiple intimate mixture
  scenario.
\newblock In {\em Proc. 6th IEEE Workshop on Hyperspectral Image and Signal
  Processing: Evolution in Remote Sensing (WHISPERS)}, 2014.

\bibitem{Saranathan16ManifoldClustering}
Arun Saranathan and Mario Parente.
\newblock Unmixing multiple intimate mixtures via a locally low-rank
  representation.
\newblock In {\em Proc. of 8th IEEE GRSS Workshop on Hyperspectral Image and
  Signal Processing: Evolution in Remote Sensing (WHISPERS)}, August, 2016.

\bibitem{Shi:2000:NCI:351581.351611}
Jianbo Shi and Jitendra Malik.
\newblock Normalized cuts and image segmentation.
\newblock {\em IEEE Trans. Pattern Anal. Mach. Intell.}, 22(8):888--905, August
  2000.

\bibitem{Souvenir05manifoldclustering}
Richard Souvenir and Robert Pless.
\newblock Manifold clustering.
\newblock In {\em In ICCV}, pages 648--653, 2005.

\bibitem{subspaceClustering}
R.~Vidal.
\newblock Subspace clustering.
\newblock {\em IEEE Signal Processing Magazine}, 28(2):52--68, March 2011.

\bibitem{wang2011local}
Yong Wang, Yuan Jiang, Yi~Wu, and Zhi-Hua Zhou.
\newblock Local and structural consistency for multi-manifold clustering.
\newblock In {\em IJCAI Proceedings-International Joint Conference on
  Artificial Intelligence}, volume~22, page 1559. Citeseer, 2011.

\bibitem{yang2012clustering}
Zhirong Yang, Tele Hao, Onur Dikmen, Xi~Chen, and Erkki Oja.
\newblock Clustering by nonnegative matrix factorization using graph random
  walk.
\newblock In {\em Advances in Neural Information Processing Systems}, pages
  1079--1087, 2012.

\end{thebibliography}
	%
	
	
	
	%
	
	\begin{IEEEbiography}[{\includegraphics[width=1in,height=1.25in,clip,keepaspectratio]{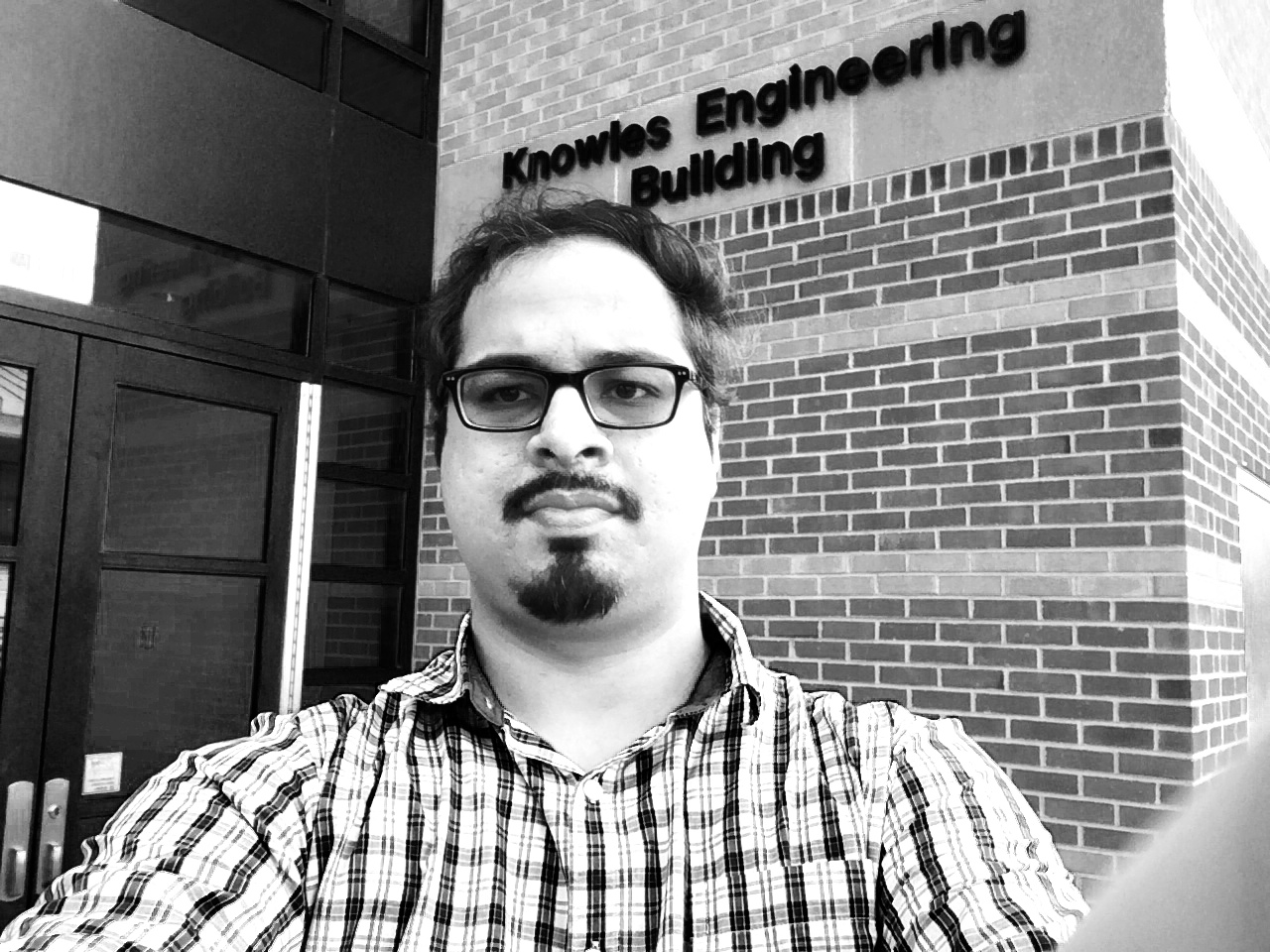}}]
		{Arun M Saranathan} received the  B.E. degree from Visvesvaraya Technological University in Belgaum, India and the M.S. degree in Electrical Engineering from the University of Massachusetts, Amherst. He is currently a Ph.D student in the Department of Electrical \& Computer Engineering at University of Massachusetts, Amherst. His interests include the use and extension of image segmentation techniques for Hyperspectral (HSI) Images and the use of manifold techniques to model the mixing seen in HSI. He has been a student member of IEEE since 2013.
	\end{IEEEbiography}
	
	\begin{IEEEbiography}
		[{\includegraphics[width=1in,height=1.25in,clip,keepaspectratio]{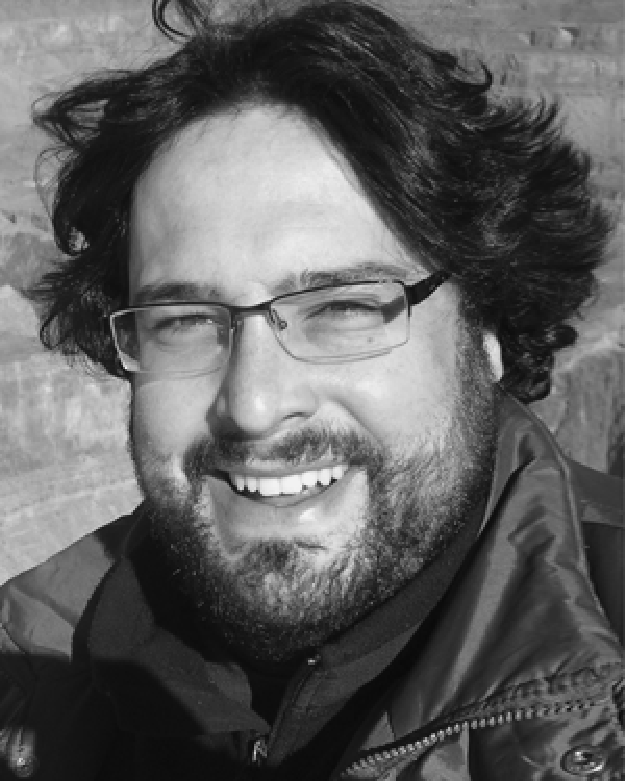}}]
		{Mario Parente} (M'05-SM'13) received the B.S. and M.S. (summa cum laude) degrees in telecommunication engineering from the University of Federico II of Naples, Italy, and the M.S. and Ph.D. degrees in electrical engineering from Stanford University, Stanford, CA. He was a Post-Doctoral Associate in the Department of Geosciences at Brown University. He is currently an Assistant Professor in the Department of Electrical and Computer Engineering at the University of Massachusetts Amherst. His research involves combining physical models and statistical techniques to address issues in remote sensing of Earth and planetary surfaces. Prof. Parente's professional interests include identification of ground composition, geomorphological feature detection and imaging spectrometer data modeling, reduction and calibration for NASA missions. He developed machine learning algorithms for the representation and processing of hyperspectral data based on statistical, geometrical and  topological models. Dr. Parente's research also involves the study of physical models of light scattering in particulate media.  Furthermore, he has developed solutions for the integration of color and hyperspectral imaging and robotics to identify scientifically significant targets for rover and orbiter-based reconnaissance. Dr. Parente has supported several scientific teams in NASA missions such as the Compact Reconnaissance Imaging Spectrometer for Mars (CRISM), the Mars Mineralogy Mapper (M3) and the Mars Science Laboratory ChemCam science teams. Dr. Parente is a principal investigator at the SETI Institute, Carl Sagan Center for Search for Life in the Universe and a member of the NASA Astrobiology Institute. Prof. Parente is serving as an Associate Editor for the IEEE Geoscience and Remote Sensing Letters.
	\end{IEEEbiography}
	
	
	

\end{document}